\documentclass[runningheads]{llncs}
\usepackage[T1]{fontenc}
\usepackage{graphicx}
\usepackage{multirow}
\usepackage{multicol}

\begin{document}
\title{Assessing Good, Bad and Ugly Arguments Generated by ChatGPT: a New Dataset, \\ its Methodology and Associated Tasks}
\titlerunning{The Good, the Bad and the Ugly Arguments Generated by ChatGPT}
 
\author{Victor Hugo Nascimento Rocha\inst{1}\orcidID{0000-0002-8984-7982} \and
Igor Cataneo Silveira\inst{2}\orcidID{0000-0003-1599-8347} \and
Paulo Pirozelli \inst{3}\orcidID{0000-0002-4714-287X} \and
Denis Deratani Mauá \inst{2}\orcidID{0000-0003-2297-6349} \and
Fabio Gagliardi Cozman \inst{1}\orcidID{0000-0003-4077-4935}}
\authorrunning{V. H. N. Rocha et al.}

\institute{Escola Politécnica 
\and Instituto de Matemática e Estatística 
\and Instituto de Estudos Avançados \\
Universidade de São Paulo\\ 
\email{\{victor.hugo.rocha@,igorcs@ime.,paulo.pirozelli.silva@, \\fgcozman@,denis.maua@\}usp.br}}
 
\maketitle       

\begin{abstract}
The recent success of Large Language Models (LLMs) has sparked concerns about their potential to spread misinformation. As a result, there is a pressing need for tools to identify ``fake arguments'' generated by such models. To create these tools, examples of texts generated by LLMs are needed. This paper introduces a methodology to obtain good, bad and ugly arguments from argumentative essays produced by ChatGPT, OpenAI's LLM. We then describe a novel dataset containing a set of diverse arguments, \textit{ArGPT}. We assess the effectiveness of our dataset and establish baselines for several argumentation-related tasks. Finally, we show that the artificially generated data relates well to human argumentation and thus is useful as a tool to train and test systems for the defined tasks.


\end{abstract}

\section{Introduction}

Recently  we have witnessed the popularization of large language models (LLMs). OpenAI's ChatGPT,\footnote{\url{https://openai.com/blog/chatgpt}} for instance, has garnered considerable media attention and public interest due to its impressive linguistic abilities and wide range of knowledge.
This has led many people to use it as an alternative to traditional search engines for gathering information. However, LLMs are not at this point reliable tools for knowledge-related tasks. 
In particular, given that their main goal is to  model  the joint probability distribution of tokens, they tend to produce 
grammatically sound 
and realistic texts that can be problematic in a number of ways, such as producing convincing arguments to justify false claims.
This is particularly worrisome considering that even hand-made datasets are getting more and more contaminated by the use of LLMs \cite{veselovsky2023artificial}.

It is that unintended aspect of LLMs that we investigate in this work.
We focus on ChatGPT, even though we expect our conclusions to be representative of any similar LLM. 
To the best of our knowledge, 
this is the first study in the literature 
with this specific aim.

Given the complexity of LLMs, such a study must be based
on empirical investigation, 
by extracting and analyzing 
a diverse set of arguments.
To this end, our first contribution is a methodology to generate arguments that are representative of ChatGPT's skills. 
Our second contribution is the dataset itself, ArGPT, a curated set of argumentative essays that
have been annotated by human experts with carefully selected labels. 
We expect this dataset to be a useful resource in our quest to understand the behavior of LLMs as regards to argumentation.  
To validate this effort and to indicate how the dataset can be
employed in practice, our third contribution 
is a well defined set of tasks and corresponding baselines; namely, Argument Quality Classification, Span Identification, Component Classification, Relation Classification, and Essay Scoring.
Finally, we show 
that our LLM argumentation dataset is sufficiently similar to human-made datasets, 
suggesting that our method is scalable for argument mining in general, enabling faster and lower-cost generation of data. 

The paper is divided as follows. Section \ref{sec:background} summarizes the needed background. Section \ref{sec:ArGPT} presents the methodology used to create the dataset. Section \ref{sec:Corpus Annotation} introduces the annotation process and statistics for ArGPT. Section \ref{sec:baselines} defines the tasks proposed for our dataset and presents their respective baselines. In Section \ref{sec:human_arg}, we show that the arguments generated with our methodology are similar to traditional, human-made Argument Mining datasets. Finally, Section \ref{sec: Conclusions} concludes and presents ideas for future work.\footnote{All code, data, and experiments for this paper are available at: \url{https://github.com/C4AI/ArGPT}.}

\section{Background}\label{sec:background}

Our 
investigation 
about the quality of arguments provided by LLMs is 
informed by two proxy tasks: Argument(ation) Mining (AM) and Automatic Essay Scoring (AES), topics
that we now briefly review.


\subsection{Argument(ation) Mining}\label{sec:AM}

Argument Mining (AM) is interested in extracting arguments, their relations and structures from natural language texts \cite{Lawrence2020}.  
It is usually divided into three different subtasks: span detection, component classification and relation classification. The first task finds the parts of an input text that are argumentative; the second classifies those components into argumentative entities (such as premises, claims, etc); and the third task classifies the relations among components (support, attack, etc) \cite{Lawrence2020,Morio2022}. Earlier methods in AM adopted algorithms based on hand-crafted features and rules, but recently  the use of contextual language models, such as   transformers, has gained prominence \cite{Hidayaturrahman2021,Mayer2020}. 

One of the major problems in AM is the heterogeneity of the existing datasets. Each corpus is based on a particular methodology and takes a different terminology to label arguments. On top of that, the size of the existing datasets tend to be limited when compared to datasets found in other NLP tasks, possibly due to the difficulty in generating and annotating arguments
\cite{Accuosto2020}. Despite this, important efforts have been undertaken to create AM datasets. 
We can mention a solid corpus about the medical field \cite{Mayer2020};
another one that tries to  solve the lack of annotated data in the scientific domain \cite{Accuosto2020};
a corpus that contains user comments about policy proposals \cite{Park2018};
a smaller corpus that is based on student argumentation \cite{Peldszus2016};
and the dataset constructed by the Project Debater by the IBM Corporation.\footnote{https://research.ibm.com/interactive/project-debater/} 
A particularly valuable resource is a corpus consisting of 402 annotated student 
essays about controversial themes \cite{StabGurevych2017}, 
where argumentative parts 
are associated with one of the labels ``MajorClaim'', ``Claim'' and ``Premise'', 
and each ``Premise'' is related to other components through an ``Attack'' or ``Support'' relation.

One significant feature that seems to be lacking in existing AM corpora is the annotation of texts containing flawed arguments. Arguments   in   datasets are commonly written by well-educated people and in contexts that have no correct answer (e.g., ``Life in a city is much better than life in the countryside'' \cite{StabGurevych2017}). This tends to result in argumentations that are almost always adequate.
A dataset with flawed arguments is a key step in building detectors for such
arguments. 

\subsection{Automatic Essay Scoring}\label{sec:AES}

Automatic Essay Scoring (AES) aims to automatically assign a grade to an essay \cite{Review-2021}.   Argumentative essays are found in student exams worldwide, such as in TOEFL,\footnote{https://www.ets.org/toefl/test-takers/ibt/about/content/writing.html} 
and play a significant role in education as they require students to engage with a subject matter; analyze different perspectives; construct logical arguments; and support their viewpoints with evidence. 
AES is a valuable tool to understand the quality of arguments. Essays with high scores are those that meet basic requirements, such as presenting a clear and concise claim, and providing strong and logical premises to support it. On the other hand, essays that receive low scores may indicate that the argumentative structure is flawed or incomplete.

AES datasets are mostly based on exams. The standard dataset, ASAP,\footnote{https://www.kaggle.com/c/asap-aes} comprises eight different prompts answered by 7th to 10th grade students, with two of them being argumentative \cite{2018-asap+}. The TOEFL dataset \cite{TOEFLdataset} is made of essays written by non-native English speakers and is sadly not publicly available. One final dataset is ArgRewrite V.2 \cite{ArgRewrite}. Its smaller size is compensated by having different levels of annotation, which are useful for a wide range of NLP tasks.

\section{Generating Argumentative Essays with ChatGPT}\label{sec:ArGPT}



As with any other empirical investigation, 
our study about the validity and quality of argumentations produced by LLMs starts with the acquisition of data.
This data must properly represent all the scenarios we may be interested in, with  examples of both good and bad argumentation containing both real and fake claims. 
In order to collect such data, 
we established some main guidelines for our interaction with ChatGPT. 
To start, we decided that the texts should be argumentative essays, as those usually involve a straightforward argumentation structure and are common both in AM and AES \cite{2018-asap+,Peldszus2016,StabGurevych2017}. 
To induce contradictions or other forms of bad argumentation, we selected only fake or contradictory claims as themes for the essays. 
We make use of ChatGPT's propensity to produce justifications, even when facing clearly wrong propositions \cite{bubeck2023sparks}. 
This way, we led ChatGPT to produce several contradictory arguments.
Finally, to speed up the process, we framed the interaction with ChatGPT as an academic setting where a student must write an essay about a topic and a professor must correct it with a second essay. 
The dataset acquisition methodology is summarized in Figure~\ref{fig:overview} and consists of the following three steps:


\begin{figure}[t]
    \centering
    \includegraphics[width=1\textwidth]{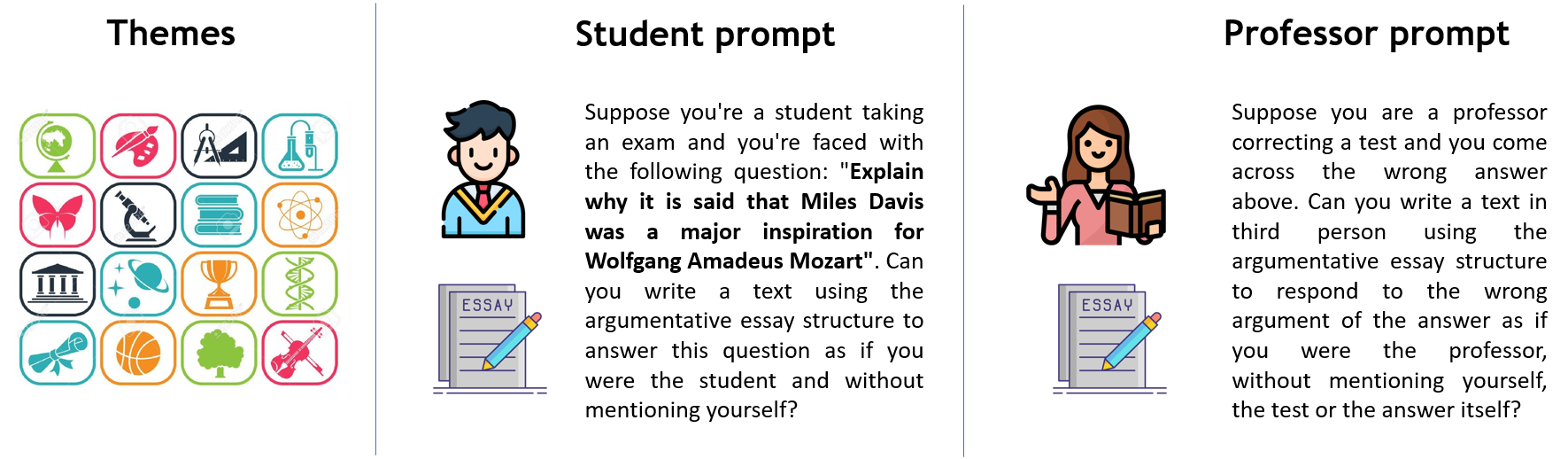}
    \caption{Generating ArGPT. We selected several themes for argumentative essays based on false or self-contradictory ideas. We gave ChatGPT a first prompt (the \textit{student} prompt), instructing it to create an argumentative essay about a selected theme. In the following round, we provided a second prompt (the \textit{professor} prompt) instructing it to write an essay correcting the student's argumentation. If the produced essays did not follow our requirements, we repeated the process.}
    \label{fig:overview}
\end{figure}

\begin{enumerate}
    \item \textbf{Themes}. The themes for the essays were chosen as to induce flawed argumentation by presenting contradictory or false ideas. 
    We selected themes from several areas, including art, history, philosophy and science. All themes were expressed in English and are available with the dataset.


    \item \textbf{Student Essay}. In order to induce the writing of the first  argumentative essay, we fed ChatGPT the prompt: \textit{Suppose you're a student taking an exam and you're faced with the following question: ``Explain {\tt [THEME]}''. Can you write a text using the argumentative essay structure to answer this question as if you were the student and without mentioning yourself?}.
    \item \textbf{Professor Essay}. In a second round of the same dialogue section, we asked ChatGPT to correct the first essay playing the role of a professor. The goal of this second interaction was to get one of two outcomes: (i) if the student had created a flawed argument, the professor would write an essay with the correct argumentation; or (ii) if the student's argument was reasonable, the professor would produce a flawed argument. The professor prompt was: \textit{Suppose you are a professor correcting a test and you come across the wrong answer above. Can you write a text in third person using the argumentative essay structure to respond to the wrong argument of the answer as if you were the professor, without mentioning yourself, the test or the answer itself?}.
\end{enumerate}

After collecting the essays, we checked whether they followed our requirements; i.e., whether the texts attended to an essay structure and tackled the proposed theme. When that was not the case, we re-prompted the themes in a new dialogue section until we got a pair of adequate argumentative essays. No theme required more than a single re-promting.
We note that prompts had to be finely engineered
through a relatively long trial and error process to produce the desired results.

\section{ArGPT: Dataset Annotation and Statistics}\label{sec:Corpus Annotation}


To produce a dataset that can in fact be used to study how ChatGPT argues, all generated arguments must be annotated in meaningful ways. To check the essays argumentative structure, we first annotated them for the three standard AM tasks (span identification, component classification, relation classification). Next, as we wanted to assess the overall quality of the written material, we also annotated texts for AES. 





To guide the AM annotations, we mostly followed the methodology by Stab \& Gurevych (2017) \cite{StabGurevych2017}.\footnote{
For the sake of space, we discuss only the differences of our methodology in respect to theirs.}
However, our approach to classifying argumentative components and relations was more minimalist
. While they differentiated three types of argumentative components (premises, claims, and major claims), we only distinguished between major claims --- the central viewpoints argued in the essays --- and premises, which supported or attacked other components, whether major claims or premises. We adopted this approach due to several reasons. Firstly, most of premises in our essays were   related to major claims. Secondly, it was more important to establish the components' relations than to categorize them. Finally, we wanted to simplify the annotation process so as to   annotate many arguments. The types of relations between arguments are two: attack, when the source component contradicted or weakened the target component, and support, when the source strengthened or justified the target.
We also had to accommodate for the idiosyncrasies of texts produced by ChatGPT. 
In argumentative essays, it is a common practice to repeat the main claim several times. 
ChatGPT's texts also followed this pattern; however, since we fed it with claims that likely went against its knowledge base, it occasionally contained contradictory major claims in the same text. 
Therefore, we established that all occurrences of a major claim in an essay should be annotated separately and that the annotator should establish the relation (attack or support) among them. 
We also pointed to annotators that some premises might not be linked (directly or indirectly) to one of the major claims, as ChatGPT sometimes presents irrelevant information or is unable to connect some of the premises to the rest of the argumentative graph.

\begin{table}[t]
    \caption{List of criteria for Automatic Essay Scoring and the corresponding scores. ``Interm.'' gives the score for each minor criteria, whereas ``Final'' gives the score for the major criteria. The AES score is an average of the four major criteria.}
    \centering
    \begin{tabular}{c|c|c|c}
     \textbf{Major Criteria} & \textbf{Minor Criteria} & \textbf{Interm.} & \textbf{Final}\\
    \hline
   \multirow{4}{*}{Structure} & Clearly states a major claim & 2.5 & \multirow{4}{*}{10}\\
   & Introduces the theme & 2.5 &\\
   & Develops the arguments throughout the text & 2.5 & \\
   &  Recapitulates the arguments in the conclusion & 2.5 &\\ \hline
   \multirow{2}{*}{Writing} & Adherence to standard language norms & 5 &  \multirow{2}{*}{10}\\ 
   & Correct use of argumentative connectives & 5 &\\ \hline
    \multirow{4}{*}{Coherence} & Adherence to the theme & 2.5 & \multirow{4}{*}{10}\\
   & No repetition of arguments & 2.5 &\\
   & No contradictions & 2.5 & \\
   &  No beating around the bush & 2.5 &\\ \hline
   Truthfulness & States true or plausible arguments & 10 & 10\\ 
    \end{tabular}
    \label{tab:scores}
\end{table}


As for the AES annotation, ArGPT's argumentative essays were evaluated with respect to quality and correctness.
To ensure consistency in our assessment, we developed a comprehensive list of criteria, modeled after evaluations from standard exams. Our list is organized into four main criteria: structure, writing, coherence, and truthfulness. \textit{Structure} measures whether the essay attends to the necessary structure of an argumentative essay; \textit{Writing} evaluates linguistic correctness; \textit{Coherence} evaluates the essay's argumentative structure; and \textit{Truthfulness} evaluates the veracity of the presented information. Each of these criteria is further divided into one or more subcriteria. As a final score, we take the arithmetic mean between these four criteria. Table \ref{tab:scores} displays the full set of criteria and the points associated with them.
Based on our methodology, 84 themes were selected, resulting in 168 argumentative essays. Each essay was generated in about 8 minutes (16 minutes per theme). As a matter of comparison, according to testing services such as TOEFL, a person takes between 30 and 50 minutes to write an essay. Hence we can see that a state-of-art conversational agent indeed operates quickly for the purposes of dataset generation.
The full annotation process was carried out by an experienced annotator specifically hired for this (not an author of this paper), who spent around 20 minutes in each essay. ArGPT is split into training (80\%), validation (10\%) and test (10\%) sets.\footnote{ChatGPT has emerged as a valuable annotation tool, often outperforming manual annotations. (e.g., \cite{gilardi2023chatgpt,törnberg2023chatgpt4}). Nonetheless, despite our best efforts, we could not teach ChatGPT to generate annotations adhering to our methodology. This limitation is reasonable, considering that even human annotators require training to perform such tasks effectively.}

\begin{table}[t]
    \caption{Statistics for ArGPT and similar AM and AES datasets. Statistics for ArGPT are for the full dataset. ``Average Number of Words'' is not reported for the AM datasets that do not provide the full text. Information on ``Structure Type'', ``Number of Components'', and ``Number of Relations'' are not available for AES datasets.}
    \centering
    \begin{tabular}{c|c|c|c|c|c|c|c|c|c}
     & ArGPT & \cite{StabGurevych2017} & \cite{Peldszus2016} & \cite{Park2018} & \cite{Mayer2020} & \cite{Accuosto2020} & \cite{2018-asap+} & \cite{TOEFLdataset} & \cite{ArgRewrite}\\
    \hline
      Corpus Type & AM+AES & AM & AM & AM & AM & AM & AES & AES & AES \\
      Structure Type  & graph & tree & tree & graph & graph & tree & - & - & - \\
      Num. texts  & 168 & 402 & 112 & 731 & 659 & 60 & 3600 & 12000 & 258\\
      Num. Components  & 2730 & 6089 & 576 & 4779 & 4198 & 353 & - & - & -\\
      Num. Relations  & 2713 & 3832 & 464 & 1353 & 2601 & 293 & - & - & -\\
      Avg. Num. Words  & 380 & 366 & - & - & 337 & 134 & 350 & 348 & 582\\
    \end{tabular}
    \label{tab:corpus_statistics}
    \vspace{-0.5cm}
\end{table}

Statistics for ArGPT are shown in Table \ref{tab:corpus_statistics}, as well as for similar AM and AES datasets (see Section \ref{sec:background}).  
The average word count per essay in ArGPT (380) is in line with the other datasets.
The statistics also show that our dataset has a higher rate of relations per component; a side effect of annotating the relations between the multiple major claims. Regarding the essay scores, they ranged from $5.25$ to $9.75$, with an average of $7.76$ and standard deviation of $0.96$.



Given annotations for AM and AES,   ChatGPT produced mostly high-quality essays, with well-distributed scores and which seem to follow an argumentative structure. Based on that, particularly the third\footnote{We excluded ``Adherence to the theme'' from this account.} and fourth criteria from Table \ref{tab:scores}, we defined a typology for the argumentation quality of the essays. Essays with a solid argumentation (high coherence) and which argued in support of a true claim were labeled ``Good''. Essays that exhibited a flawed argumentation were labeled ``Bad'', regardless of whether they defended something that was true or not. Finally, texts that did a quite good job in defending false claims were labeled ``Ugly''. We consider this to be a particularly dangerous class of arguments, since it has the potential to successfully convince people to accept a false claim. Out of the 168 essays, 81 were categorized as ``Bad'', 50 were labeled as ``Good'', and 37 were marked as ``Ugly'' (the training, validation and test sets all closely follow that proportion of labels). These numbers show that our methodology, which involved simulating a student-teacher interaction and prompting false claims, was successful in producing diverse arguments.

\section{Using ArGPT: Supported Tasks and Their Baselines}\label{sec:baselines}

The purpose of building a LLM-based argumentation dataset was to obtain a resource for   detecting and evaluating arguments produced by these models. These processes range from identifying the individual components in an argument up to assessing its global structure. For this reason, we defined five tasks that together account for the whole process of argument identification and classification. The first is a novel task of argument quality evaluation, the next three are   typical within AM \cite{Morio2022},  and the last is simply AES:

\begin{enumerate}
    \item \textit{Argumentation Quality Evaluation}: In this task, a model receives an essay as input and has to predict the quality of the argumentation. There are three possible labels: {\tt Good}, {\tt Bad}, and {\tt Ugly}.
    \item \textit{Span Identification}: Given an essay, a model has to predict the correct BIO tag (Beginning, Inside, Outside) for each token in the text.
     \item \textit{Component Classification}: In this task, components are classified as {\tt Premise} or {\tt Major Claim}. The input is a concatenation of the component and the text (``{\tt component + [SEP] + text}''), and the output is one of the two labels. We provide the full text together with the component because the role of an argumentative component depends on the context.
    \item \textit{Relation Classification}: Given two argument components, a model needs to determine their relation, using one of three possible labels: {\tt Attack}, {\tt Support}, and {\tt None}. The input for this task is a concatenation of the two components (``{\tt source-component + [SEP] + target-component}'').
    \item \textit{Essay Scoring}: This is modeled as a regression task. A model receives an argumentative essay as an input and must predict its score.
\end{enumerate}

For each of these tasks, we establish a couple of baselines, based on two standard transformer-based architectures: BERT \cite{BERT2018} and RoBERTa \cite{RoBERTa2019} base. After some initial tests, all the models were trained using the same hyperparameter configurations: 15 epochs, batch size of 8 and learning rate of 2e-5. This decision, along with the one to model the tasks as described above, was taken for simplicity's sake, since our goal in this paper is to study our dataset, with the construction of a more complex model being the subject of future work.

\subsection{Evaluation Metrics}

To evaluate the first four tasks, we utilize F1-score micro and macro as our metrics. To make their utilization in AM tasks clear, we follow the definitions by Morio et al.\ (2022) \cite{Morio2022}. The first task (Span Identification) predicts a set of spans, defined as a pair $(s, e)$, where $s$ represents the span's start token and $e$ the end token. The second task (Component Classification) predicts a set of components $(s, e, c)$, where $s$ and $e$ have the same meaning as before and $c$ is the component's label. Lastly, the third task (Relation Classification) predicts a set of relations, with each one defined as $(s_{src}, e_{src}, s_{tgt}, e_{tgt}, r)$, where $s_{src}$ and $e_{src}$ represent the source component's span, $s_{tgt}$ and $e_{tgt}$ the target's span and $r$ the relation label. Given this, the evaluation metrics for the AM tasks can be defined. Consider $\mathcal{G}_{task}$ the set of gold outputs of a certain AM task and $\mathcal{S}_{task}$ the set of system outputs for the same task. The precision metric is defined as $P = |\mathcal{G}_{task} \cap \mathcal{S}_{task}| / |\mathcal{S}_{task}|$, the recall metric is $R = |\mathcal{G}_{task} \cap \mathcal{S}_{task}| / |\mathcal{G}_{task}|$ and, finally, the F1-score is $F = 2PR / P + R$.

For the AES task, we utilize two metrics. The first is the Quadratic Weighted Kappa (QWK), which measures the degree of agreement between two graders. The QWK score ranges from -1 to 1, where 0 indicates random agreement, 1 indicates perfect agreement, and -1 indicates perfect disagreement. Although the effectiveness of QWK is debated \cite{Evaluation-Metrics}, it is widely used in the AES field due to the influence of the ASAP dataset. The second metric employed in this task is the Mean Squared Error (MSE), usually used for regression tasks.


\subsection{Results and Discussion}

\begin{table}[t]
    \caption{Baseline results for ArGPT's test set. The first two rows report the values for models trained in individual tasks (no error propagation). The last row shows the result achieved in each step of an end-to-end AM pipeline allowing the error to propagate and using the RoBERTa models trained for each individual task.
    }
    \centering
    \begin{tabular}{c|c|c|c|c|c|c|c|c|c}
		\multirow{9}{*}{} & \multicolumn{2}{c|}{Arg. Qual.} & {Span} & \multicolumn{2}{c|}{Component} & \multicolumn{2}{c|}{Relation} &  \multicolumn{2}{c}{AES}
		\\
    \cline{2-10}
  {} & {F1} & {Macro} & {F1} & {F1} & {Macro} & {F1} & {Macro} & {MSE} & {QWK}\\
	\cline{1-10}
		   BERT & \textbf{56.25\%} & 40.40\%& 50.56\% &  90.87\% & 78.74\% & \textbf{93.91\%} & 36.81\% & 0.58 & 0.53\\
	\hline
        RoBERTa   &\textbf{56.25\%} & \textbf{54.97\%} & \textbf{77.35\%} &  \textbf{92.34\%} & \textbf{81.67\%} & 92.36\% & \textbf{44.18\%} & \textbf{0.44} & \textbf{0.65}\\
    \hline
    \hline
        Pipeline & - & - & 77.35\% & 71.07\% & 71.46\% & 52.89\% & 37.58\% & - & -\\ 
    \end{tabular}
    \label{tab:first_results}
    \vspace{-0.5cm}
\end{table}

Table \ref{tab:first_results} illustrates the results achieved in the different tasks for ArGPT's test set. The first two lines report the values for BERT and RoBERTa models in the individual tasks, with RoBERTa performing better in all tasks. 

The RoBERTa model achieved an F1-macro score of 54.97\% in the Argument Quality task. This is evidence that, to some extent, our models were able to differentiate the good, bad and ugly arguments produced by ChatGPT. It should be noted however that while our models do contain some form of knowledge in their parametrization, their ability to fact-check claims is limited, since they do not have access to up-to-date knowledge databases. This shortcoming is part of the reason why the results were not better in certain cases.

For AM, in addition to the individual tasks, we simulated an end-to-end pipeline, i.e. from raw texts to the argumentation graphs, in which errors are able to propagate between tasks. For each part of the pipeline, we took the model that performed better in the standalone task (shown in bold in  Table \ref{tab:first_results}), which was RoBERTa in all cases. The pipeline starts with the trained span identification model. The output from this model (the identified components), instead of the gold-standard spans, are then separately passed to the component classifier and to the relation classifier. The ``Pipeline'' row in Table \ref{tab:first_results} reports the values achieved by the simulation. As expected, the performances for both the component and relation classifiers were worse when the error was allowed to propagate. Nonetheless, the results were comparable to those found in the literature for similar datasets, both with and without error propagation \cite{Hidayaturrahman2021,Morio2022}. We also note that the results obtained by our models in AES were similar to others in the literature \cite{2018-asap+}.




\section{The Connection with Human Argumentation}
\label{sec:human_arg}

\begin{table}[t]
    \caption{The first three rows bring the results for the Stab \& Gurevych's dataset \cite{StabGurevych2017}, for three models: i) a pretrained model (no fine-tuning); ii) a model fine-tuned on this dataset \cite{Morio2022}; iii) and the best models trained on ArGPT. The last row presents the results of a model trained on Stab \& Gurevych's dataset and tested on ArGPT.}
    \centering
    \begin{tabular}{c|c|c|c|c|c}
		\multirow{5}{*}{} & {Span} & \multicolumn{2}{c|}{Component} & \multicolumn{2}{c}{Relation} \\
    \cline{2-6}
  {} & {F1} & {F1} & {Macro} & {F1} & {Macro} \\
	\cline{1-6}
        Pretrained Model   &  0\% &  12.09\% & 10.78\% & 0.02\% & 0.03\%\\
    \hline
        Recent Result \cite{Morio2022}   & 85.20\% &  87.68\% & 80.37\% & 66.91\% & 55.84\%\\
    \hline
    Our Best Model & 53.48\% &  91.63\% & 76.31\% & 92.98\% & 34.11\%\\
	\hline
    \hline
    Model trained in \cite{StabGurevych2017} & 48.51\% &  87.23\% & 55.84\% & 86.85\% & 34.26\%\\
    \end{tabular}
    \label{tab:second_results}
    \vspace{-0.5cm}
\end{table}

The arguments generated by ChatGPT resemble those found in human augmentation. 
Yet, one might wonder whether ArGPT dataset is useful for performing AM for human-generated texts.


To verify that hypothesis we designed two experiments. In the first, we tested whether models trained on ArGPT generalize to the dataset by Stab \& Gurevych (2017) \cite{StabGurevych2017}. The rationale is that if a model trained in ArGPT performs well in human-generated essays, then we can leverage ChatGPT for cheaper   training of AM systems. 
For consistency, we modified the target dataset annotations by changing ``Claim'' labels to ``Premise'', and applying our changes to ``Major Claims''. The   results,   in Table \ref{tab:second_results}, are compared to both a pretrained model (RoBERTa with no fine-tuning) and the results by Morio et al.\ (2022) \cite{Morio2022} (even if this comparison is limited due to methodological changes). Compared to a pretrained
model, our model did learn to solve tasks related to human
argumentation and very decently performed them. And, although the results achieved by our model were far from
state of the art approaches, we must remember that it was not fine-tuned in the
target dataset and still got some decent F1-metrics.

For the second experiment, we took  the opposite route. We trained a model (RoBERTa with the same hyperparameters as before) in Stab \& Gurevych's dataset and tested it on ArGPT's test set. 
The results, shown in the last row of Table \ref{tab:second_results}, were similar. 

Together, the two experiments strongly suggest that ArGPT (and our methodology in general) can be effectively used to learn AM models that emulate human datasets, with two clear advantages. 
First, text generation by ChatGPT is significantly faster and cheaper than by (paid) humans.
Secondly, one can produce texts containing features that are  difficult for non-expert human annotators to intentionally produce, such as contradictions, defenses of false claims and so on. 
Those advantages can mitigate the current poverty of annotated data in AM.

 


\section{Conclusions and Future Work}
\label{sec: Conclusions}

We have developed a number of techniques to assess the ability of LLMs to argue, in 
particular looking at OpenAI's ChatGPT; we thus introduced a new dataset, ArGPT,
consisting of essays that argue over false claims following a teacher-professor dialogue simulation. The dataset is annotated in a number of structural and qualitative ways, following a purposefully-developed methodology. We also defined five tasks related to argumentation and provided baselines for them. Our results indicate that it is possible to differentiate, at least to some extent, between good, bad and ugly arguments produced by ChatGPT. 
Additionally, we have shown that our method is also useful to support AM and AES applications.
Indeed, ArGPT and its associated methodology can be used to (i) develop systems capable of identifying problematic argumentation generated by LLMs, (ii) train and evaluate AM and AES systems, and (iii) speed up the generation of data for these and related tasks.

As our methods should be applicable to any LLM, it is still necessary to actually investigate how the arguments produced by other LLMs differ from the ones created by ChatGPT. This is particularly important as new open-source LLMs are now released almost every day \cite{gpt4all,alpaca}. Contrary to ChatGPT, not all LLMs  have carefully curated data or enforce content barriers; thus, our methods can be even more valuable in those cases. 

In future work we will increase the size of ArGPT and improve baseline models for all tasks. We plan to create a larger   ArGPT by an  improved annotation process, where two annotators will combine their annotations into a consensus. To create more robust baselines, we intend to explore models that can deal with structured knowledge or that can incorporate the argumentation structures extracted by the AM tasks. 
We also plan to deal with pieces other than essays, to make our dataset more diverse and applicable to a broader range of research.

\subsubsection*{Acknowledgements} 

This work was carried out at the Center for Artificial Intelligence (C4AI-USP),  with support by  FAPESP grant 2019/07665-4  and by the IBM Corporation. Victor Hugo is partially supported by the Coordenação de Aperfeiçoamento de Pessoal de Nível Superior - Brasil (CAPES) grant 88887.616392/2021-00. 
Paulo is supported by the FAPESP grant 2019/26762-0.  
Denis is partially supported by grants FAPESP \#2022/02937-9 and CNPq \#305136/2022-4.
Fabio is partially supported by CNPq \#305753/2022-3. 
Igor is partially supported by CAPES grant 88887.635492/2021-00. 
We acknowledge support by CAPES - Finance Code 001. 
\vspace{-0.1    cm}
 
\bibliographystyle{splncs04}
\bibliography{bibliography} 

\begin{thebibliography}{10}
\providecommand{\url}[1]{\texttt{#1}}
\providecommand{\urlprefix}{URL }
\providecommand{\doi}[1]{https://doi.org/#1}

\bibitem{Accuosto2020}
Accuosto, P., Saggion, H.: Mining arguments in scientific abstracts with
  discourse-level embeddings. Data \& Knowledge Engineering  \textbf{129},
  101840 (2020)

\bibitem{gpt4all}
Anand, Y., Nussbaum, Z., Duderstadt, B., Schmidt, B., Mulyar, A.: Gpt4all:
  Training an assistant-style chatbot with large scale data distillation from
  gpt-3.5-turbo. \url{https://github.com/nomic-ai/gpt4all} (2023)

\bibitem{TOEFLdataset}
Blanchard, D., Tetreault, J., Higgins, D., Cahill, A., Chodorow, M.: Toefl11: A
  corpus of non-native english. ETS Research Report Series  \textbf{2013},
  i--15 (2013)

\bibitem{bubeck2023sparks}
Bubeck, S., Chandrasekaran, V., Eldan, R., Gehrke, J., Horvitz, E., Kamar, E.,
  Lee, P., Lee, Y.T., Li, Y., Lundberg, S., Nori, H., Palangi, H., Ribeiro,
  M.T., Zhang, Y.: Sparks of artificial general intelligence: Early experiments
  with gpt-4 (2023)

\bibitem{BERT2018}
Devlin, J., Chang, M.W., Lee, K., Toutanova, K.: {BERT: Pre-training of deep
  bidirectional transformers for language understanding}. In: NAACL HLT 2019 -
  2019 Conference of the North American Chapter of the Association for
  Computational Linguistics: Human Language Technologies - Proceedings of the
  Conference. vol.~1, pp. 4171--4186. Association for Computational Linguistics
  (ACL) (2019)

\bibitem{gilardi2023chatgpt}
Gilardi, F., Alizadeh, M., Kubli, M.: Chatgpt outperforms crowd-workers for
  text-annotation tasks (2023)

\bibitem{Hidayaturrahman2021}
Hidayaturrahman, Dave, E., Suhartono, D., Arymurthy, A.M.: {Enhancing
  argumentation component classification using contextual language model}.
  Journal of Big Data  \textbf{8}(1), ~103 (dec 2021)

\bibitem{ArgRewrite}
Kashefi, O., Afrin, T., Dale, M., Olshefski, C., Godley, A., Litman, D., Hwa,
  R.: {ArgRewrite} v.2: an annotated argumentative revisions corpus. Language
  Resources and Evaluation  \textbf{56}(3),  881--915 (jan 2022)

\bibitem{Review-2021}
Lagakis, P., Demetriadis, S.: Automated essay scoring: A review of the field.
  In: 2021 International Conference on Computer, Information and
  Telecommunication Systems (CITS). pp.~1--6 (2021)

\bibitem{Lawrence2020}
Lawrence, J., Reed, C.: {Argument Mining: A Survey}. Computational Linguistics
  \textbf{45}(4),  765--818 (jan 2020)

\bibitem{RoBERTa2019}
Liu, Y., Ott, M., Goyal, N., Du, J., Joshi, M., Chen, D., Levy, O., Lewis, M.,
  Zettlemoyer, L., Stoyanov, V.: Roberta: {A} robustly optimized {BERT}
  pretraining approach. CoRR  \textbf{abs/1907.11692} (2019)

\bibitem{2018-asap+}
Mathias, S., Bhattacharyya, P.: {ASAP}++: Enriching the {ASAP} automated essay
  grading dataset with essay attribute scores. In: International Conference on
  Language Resources and Evaluation ({LREC} 2018). European Language Resources
  Association (ELRA), Miyazaki, Japan (May 2018)

\bibitem{Mayer2020}
Mayer, T., Cabrio, E., Villata, S.: {Transformer-based argument mining for
  healthcare applications}. In: Frontiers in Artificial Intelligence and
  Applications. vol.~325, pp. 2108--2115. IOS Press BV (aug 2020)

\bibitem{Morio2022}
Morio, G., Ozaki, H., Morishita, T., Yanai, K.: {End-to-end Argument Mining
  with Cross-corpora Multi-task Learning}. Transactions of the Association for
  Computational Linguistics  \textbf{10},  639--658 (may 2022).
  \doi{10.1162/tacl_a_00481}

\bibitem{Park2018}
Park, J., Cardie, C.: A corpus of e{R}ulemaking user comments for measuring
  evaluability of arguments. In: International Conference on Language Resources
  and Evaluation ({LREC} 2018). European Language Resources Association (ELRA),
  Miyazaki, Japan (May 2018)

\bibitem{Peldszus2016}
Peldszus, A., Stede, M.: {An Annotated Corpus of Argumentative Microtexts}.
  European Conference on Argumentation (ECA'16) pp. 801--816 (2016)

\bibitem{StabGurevych2017}
Stab, C., Gurevych, I.: Parsing argumentation structures in persuasive essays.
  Computational Linguistics  \textbf{43}(3),  619--659 (09 2017)

\bibitem{alpaca}
Taori, R., Gulrajani, I., Zhang, T., Dubois, Y., Li, X., Guestrin, C., Liang,
  P., Hashimoto, T.B.: Stanford {Alpaca}: An instruction-following {LLaMA}
  model (2023)

\bibitem{törnberg2023chatgpt4}
Törnberg, P.: Chatgpt-4 outperforms experts and crowd workers in annotating
  political twitter messages with zero-shot learning (2023)

\bibitem{veselovsky2023artificial}
Veselovsky, V., Ribeiro, M.H., West, R.: Artificial artificial artificial
  intelligence: Crowd workers widely use large language models for text
  production tasks (2023)

\bibitem{Evaluation-Metrics}
Yannakoudakis, H., Cummins, R.: Evaluating the performance of automated text
  scoring systems. In: Workshop on Innovative Use of {NLP} for Building
  Educational Applications. pp. 213--223. Association for Computational
  Linguistics, Denver, Colorado (Jun 2015)

\end{thebibliography}

\end{document}